\pdfoutput=1
\documentclass[11pt]{article}

\usepackage[final]{acl} % 使用review模式

\usepackage{times}
\usepackage{pifont}

\usepackage{latexsym}

\usepackage[T1]{fontenc}
\usepackage[utf8]{inputenc}

\usepackage{microtype}

\usepackage{inconsolata}

\usepackage{graphicx}

\usepackage{amsmath}
\usepackage{cleveref}

\usepackage{amssymb}
\usepackage{multirow}
\usepackage{arydshln}
\usepackage{graphicx}
\usepackage{wrapfig}
\usepackage{colortbl}
\usepackage{booktabs}
\usepackage{makecell}

\usepackage{diagbox}
\usepackage{float}

\RequirePackage{xspace}
\makeatletter
\DeclareRobustCommand\onedot{\futurelet\@let@token\@onedot}
\def\@onedot{\ifx\@let@token.\else.\null\fi\xspace}

\title{
 
ROGLE: Robust Global-Local Alignment with Automated Region Supervision for Text-Based Person Search}

\author{
    \textbf{Chaodong Jia\textsuperscript{*}},
    \textbf{Zequn Xie\textsuperscript{*}},
    \textbf{Xibei Jia\textsuperscript{*}},
    \textbf{Sihang Cai\textsuperscript{}},\\
    \textbf{Shulei Wang\textsuperscript{}},
    \textbf{Tao Jin\textsuperscript{} \thanks{Corresponding Author.}} \\
    \\
    \normalsize \textsuperscript{} 
    \\
    \\
    \normalsize \\ %%% 请将下面的邮箱列表替换为所有六位作者的真实邮箱 %%%
    \normalsize Correspondence: zqxie@zju.edu.cn
}

\begin{document}
\maketitle

\begin{abstract}
Text-Based Person Search (TBPS) aims to retrieve pedestrian images using natural language queries. However, existing TBPS models, especially those based on CLIP, struggle with fine-grained understanding due to global representational bias and semantic sparsity inherited from training on short captions. This results in weak fine-grained alignment, exacerbated by the scarcity of region-level annotations. To address this, we propose \textbf{ROGLE} (Robust Global-Local Embedding), a unified framework that overcomes reliance on costly manual annotations through an automated Region-to-Sentence Matching (RSM) strategy. RSM automatically mines pseudo region-sentence pairs for scalable fine-grained supervision. Furthermore, ROGLE employs a multi-granular learning strategy that fuses global contrastive learning with region-level local alignment. We also introduce the \textbf{P-VLG Benchmark}, a large-scale dataset constructed by curating and enriching images from established public benchmarks . It features over 100,000 annotated regions and rich long-form captions, making it the first TBPS benchmark to support both global and local assessment protocols. Extensive experiments show that ROGLE significantly outperforms existing approaches, particularly on challenging long-form queries. Code and the P-VLG benchmark will be made publicly available.
\end{abstract}

\section{Introduction}

Text-Based Person Search (TBPS) is a challenging yet important retrieval task that aims to locate images of a target individual from a large gallery based solely on natural language descriptions~\cite{li2017person}. Unlike traditional person re-identification methods that rely on example images~\cite{he2021transreid, luo2019bag}, TBPS better models practical scenarios such as surveillance analysis or missing-person investigations, where no visual examples are available and verbal witness reports serve as the only source of information. Consequently, TBPS plays a vital role in enabling retrieval systems to operate effectively in open-world and zero-shot settings.

\begin{figure}[t]
    \centering
    \includegraphics[width=0.5\textwidth]{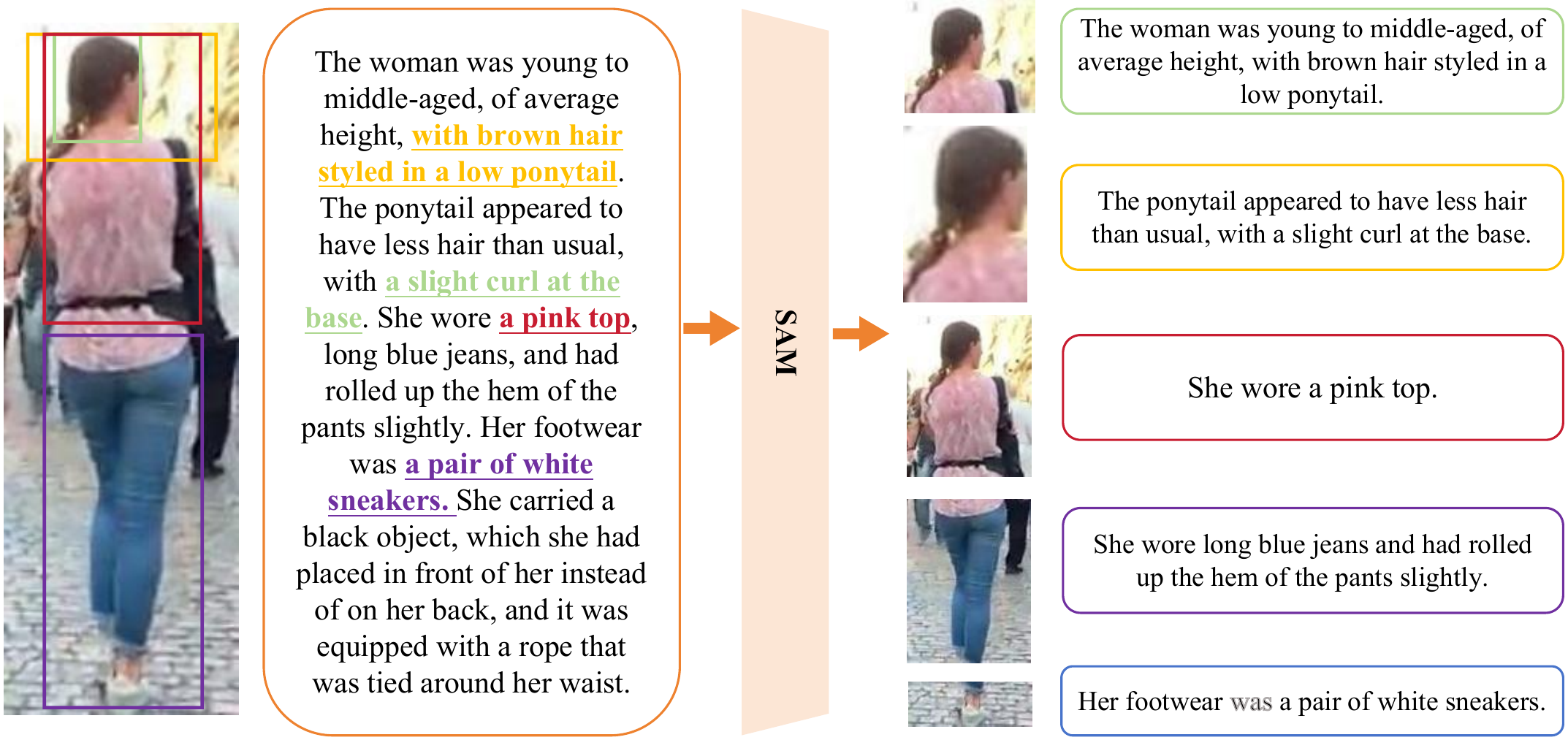}
    \caption{Visualization of our Region-to-Sentence Matching (RSM) pipeline. The original person image (left) is paired with a long descriptive caption that is first decomposed into independent sentences. We apply SAM to segment the image into semantic regions (colored boxes) and compute cross-modal similarity between each region and each sentence. The highest-scoring sentence for each region is selected, producing five local region-sentence pairs on the right, corresponding to hairstyle, clothing, and footwear.}
    \label{fig:introduction}
\end{figure}

Recent advances in TBPS have been largely driven by benchmark datasets such as CUHK-PEDES~\cite{li2017person}, ICFG-PEDES~\cite{ding2021semantically}, and RSTPReid~\cite{zhu2021dssl}. Despite this progress, a key limitation persists: most existing systems, typically built upon large vision-language models such as CLIP~\cite{IRRA,xie2025dynamic,qin2024noisy}, struggle with fine-grained understanding. Since these models are pre-trained on short, generic captions, they capture global semantics while neglecting local details. Consequently, training and evaluation datasets contain mainly brief, attribute-based annotations, offering only coarse-grained supervision that restricts a model's ability to distinguish visually similar individuals or interpret complex queries. Moreover, natural language descriptions often include abstract relations or localized cues that do not directly correspond to visual regions, making fine-grained alignment challenging. The lack of region-level annotations further constrains learning meaningful part-level correspondences.

To address the data-level bottleneck, we construct a large-scale benchmark named P-VLG using a novel Region-to-Sentence Matching (RSM) module. RSM decomposes long-form textual descriptions into atomic sentences and aligns them with semantically coherent image regions extracted by SAM~\cite{SAM}. By leveraging CLIP~\cite{clip} embeddings to compute cross-modal similarity, RSM automatically mines pseudo region-sentence pairs, providing fine-grained supervision without manual annotation. The P-VLG benchmark overcomes the limitations of prior TBPS datasets, which often lack dense grounding and favor short captions. It contains 6,801 training identities, 48,485 images, and 68,990 enriched captions, with a test set of 400 unseen identities and 6,658 image-caption pairs. The validation set provides over 100,000 region-caption alignments, enabling evaluation of both global retrieval and region-level grounding.

Based on these pseudo-labels, ROGLE adopts a dual-branch training paradigm. The first branch performs global contrastive learning to align whole-image and full-caption embeddings, while the second branch provides fine-grained supervision through multi-granular learning that integrates global and region-level local alignment. This design encourages the model to learn both contextualized global semantics and discriminative local features, essential for resolving ambiguous or compositional queries. In addition, a reliability-aware alignment calibration mechanism enhances robustness against noisy correspondences in large-scale data. Our contributions are summarized as follows:
\begin{itemize}
    \item We propose ROGLE, a unified global--local framework tackling semantic sparsity and misalignment. It incorporates an automated Region-to-Sentence Matching (RSM) module to mine pseudo-alignments from long captions, enabling fine-grained supervision without manual annotation.

    \item We introduce the P-VLG Benchmark, the first dataset supporting multi-granular TBPS evaluation. It features enriched  descriptions and over 100,000 region-level annotations to assess both holistic and localized alignment.

    \item We design a multi-granular learning strategy synergizing global contrastive learning with region-level alignment. Extensive experiments demonstrate superior performance on challenging benchmarks, particularly for detailed and compositional queries.
\end{itemize}

\begin{figure*}[t]
    \centering
    \includegraphics[width=1\textwidth]{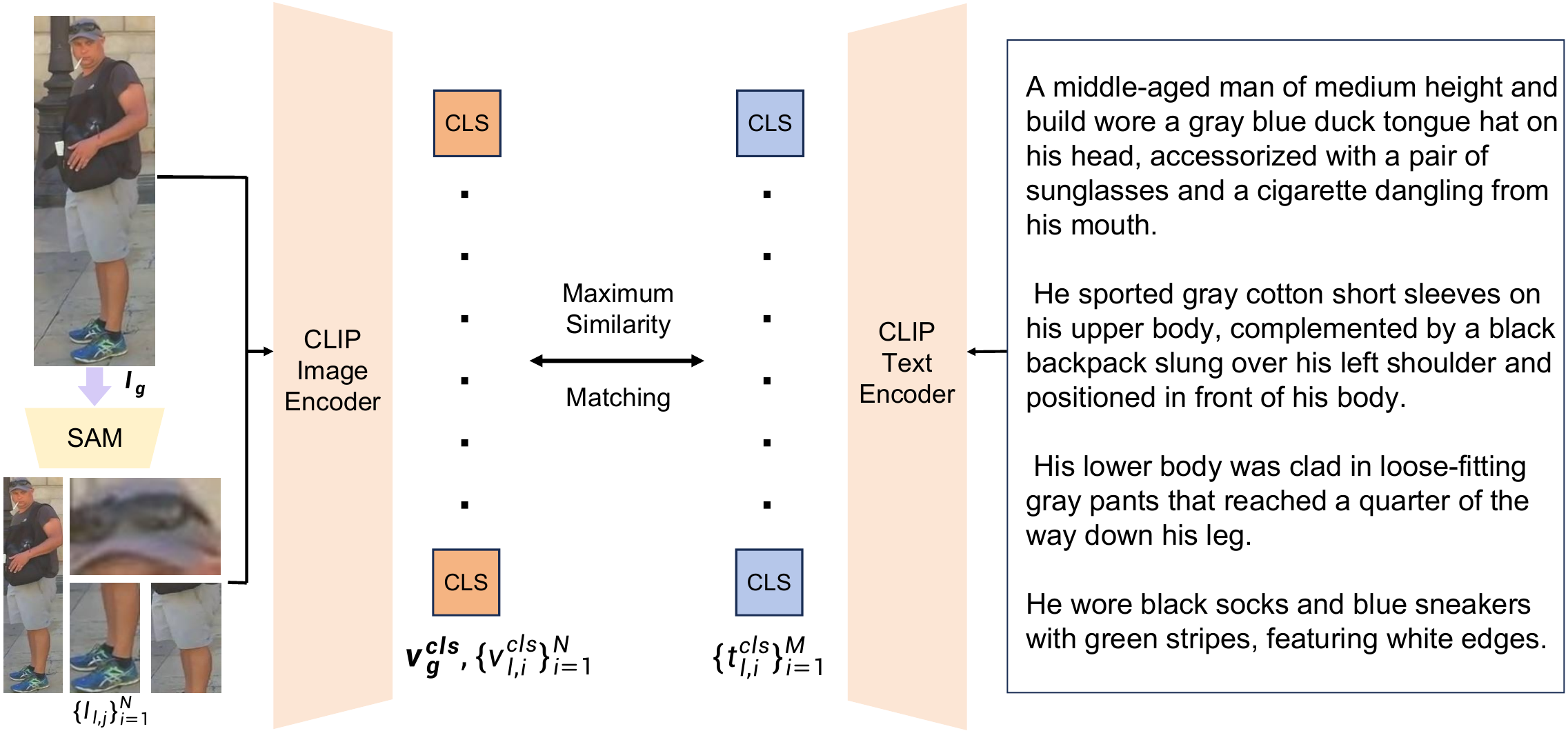}
    \caption{Visualization of the Region-to-Sentence Matching process in our P-VLG Benchmark. Given a pedestrian image and its long-form caption, SAM is used to generate region proposals $\{I_{l,i}\}_{i=1}^N$ from the global image $I_g$. All image regions and textual sentences $\{t_{l,j}^{cls}\}_{j=1}^M$ are encoded with CLIP, and cross-modal similarity is computed for all region-sentence pairs. The highest-scoring pair is then selected to create pseudo-aligned fine-grained supervision.}
    \label{fig:pipeline}
\end{figure*}
\section{Related Work}

\begin{figure*}[t]
    \centering
    \includegraphics[width=1.07\textwidth]{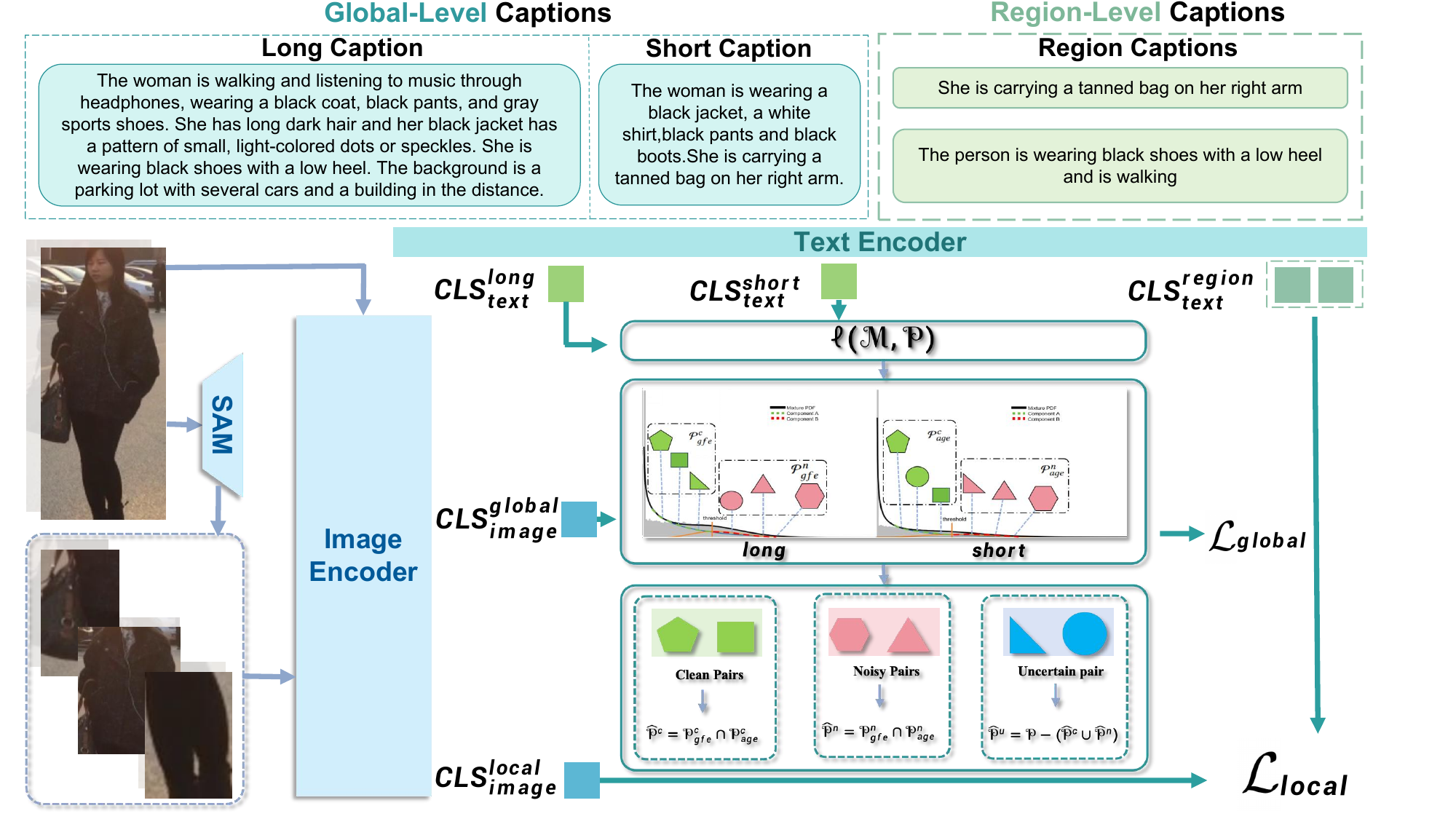} 
    \caption{Overview of ROGLE. SAM segments the input image into local regions. We employ three textual supervisions: global long, concise short, and region-level captions. Dual encoders extract text ($\text{CLS}^{\text{long}}_{\text{text}}, \text{CLS}^{\text{short}}_{\text{text}}, \text{CLS}^{\text{region}}_{\text{text}}$) and image ($\text{CLS}^{\text{global}}_{\text{image}}, \text{CLS}^{\text{local}}_{\text{image}}$) representations. A probabilistic module estimates pair reliability (clean/noisy/uncertain) to guide global ($\mathcal{L}_{\text{global}}$) and local ($\mathcal{L}_{\text{local}}$) alignment.}
    \label{fig:3}
\end{figure*}

\subsection{Text-Based Person Search}
Text-Based Person Search (TBPS) is a practical task that retrieves pedestrian images from large-scale galleries using free-form textual queries~\cite{li2017person}. Early methods mainly relied on global contrastive learning to align whole-image and full-text representations~\cite{zhang2018deep, shu2022see}. Recent advances have explored more sophisticated frameworks, such as agentic approaches for document-based retrieval~\cite{shu2026agenticretrievalaugmentedgenerationfinancial} and interaction-driven generation models~\cite{xie2025chat}. Additionally, newer studies have focused on query enhancement and context-aware representations~\cite{xie2026conquer}, as well as hierarchical visual perception to improve retrieval robustness~\cite{xie2026delving}. While effective for coarse semantic matching, global-only supervision struggles to capture subtle visual details, such as accessories, hairstyles, or multi-layered clothing, which are critical for distinguishing identities. This limitation motivates approaches that integrate both holistic and fine-grained alignment.

\subsection{Fine-Grained Alignment}
To achieve finer alignment, existing strategies fall into two categories. Manual or part-based localization methods~\cite{gao2021contextual, wang2021text, shao2022learning} align phrases with corresponding image regions but rely heavily on labor-intensive annotations or specialized part-parsing modules, limiting scalability and generalization~\cite{zuo2024ufinebench}. Detector-based supervision approaches, such as ViTAA~\cite{wang2020vitaa}, CLOC~\cite{chen2024contrastive}, and OWLv2~\cite{minderer2023scaling}, perform localized alignment via attribute modeling or open-vocabulary detection. Furthermore, research has extended into specialized domains, such as anomaly search utilizing cascade frameworks~\cite{xie2026bridgingposesemanticgapcascade} and video-based representation learning~\cite{xie2026hvd, Feng_Cai_Xie_Wu_Jin_2026}. While these methods achieve strong spatial correspondence, they often require large-scale pretraining and high computational cost. To overcome these challenges, we introduce the automatic Region-to-Sentence Matching (RSM) mechanism, which leverages pretrained CLIP features and segmentation masks from SAM~\cite{kirillov2023segment} to generate high-quality local supervision without manual annotations. This annotation-free, global-local alignment framework supports scalable fine-tuning while maintaining strong performance on TBPS tasks.
\section{Method}
\label{sec:method}

We present ROGLE, a unified framework designed to enhance fine-grained understanding in TBPS through a global-local learning paradigm. Unlike prior methods that handle global and local representations separately or assume perfect modality alignment, ROGLE integrates scalable pseudo-pair generation with hierarchical training. This design leverages both the descriptive richness of textual annotations and the structural complexity of pedestrian images, enabling robust alignment at multiple granularities.

\subsection{Problem Statement}

\label{subsec:problem}

Text-Based Person Search (TBPS) aims to retrieve the most relevant pedestrian image from a gallery given a free-form textual description. Let the gallery be 
$
\mathcal{V} = \{ \mathcal{I}_i, y^p_i, y^v_i \}_{i=1}^{N_v},
$
where $\mathcal{I}_i$ denotes the $i$-th image, $y^p_i \in \{1, \dots, C\}$ is the person identity, and $y^v_i \in \{1, \dots, N_v\}$ is the image-specific index. The corresponding textual description set is 
$
\mathcal{T} = \{ \mathcal{T}_j, y^v_j \}_{j=1}^{N_t},
$
where $\mathcal{T}_j$ describes the pedestrian in image $\mathcal{I}_j$.

During training, the dataset consists of cross-modal pairs 
$
\mathcal{P} = \{ (\mathcal{I}_i, \mathcal{T}_j), l_{ij} \}_{i=1}^{N},
$
where $l_{ij} \in \{0,1\}$ indicates whether the image and text correspond to the same person. A positive pair ideally satisfies both identity and image alignment, i.e., $y^p_i = y^p_j$ and $y^v_i = y^v_j$. However, real-world datasets often contain noisy correspondences due to annotation errors, which can introduce false positives and degrade retrieval performance. TBPS therefore requires models capable of handling both global semantic alignment and fine-grained visual distinctions, such as clothing details, accessories, and localized attributes, while being robust to noisy supervision.

\subsection{Generating the P-VLG Benchmark}

\subsubsection{Region-to-Sentence Matching}
\label{subsec:rsm}

To enable fine-grained supervision without manual annotation, we introduce the \textbf{Region-to-Sentence Matching (RSM)} module. Given a pedestrian image $\mathcal{I}_g$ and its associated long-form caption $\mathcal{T}_g$, RSM aims to generate pseudo region–sentence pairs $(\mathcal{I}_l, \mathcal{T}_l)$ that capture localized visual-text correspondences. The process consists of two main steps:

\textbf{1. Sentence decomposition:} The long-form caption $\mathcal{T}_g$ is segmented into $M$ atomic sentences $\{\mathcal{T}_l^{(j)}\}_{j=1}^{M}$ using the spaCy library~\cite{honnibal2017spacy}. Each sentence typically describes a specific visual attribute, such as clothing, hairstyle, or accessories.

\textbf{2. Region localization:} We apply the Segment Anything Model (SAM)~\cite{kirillov2023segment} with a ViT-H backbone in automatic mode to obtain $N$ semantic masks $\{\mathcal{M}^{(i)}\}_{i=1}^{N}$ from $\mathcal{I}_g$. Each mask is converted into a bounding box $\mathcal{B}^{(i)}$ to crop local image regions $\{\mathcal{I}_l^{(i)}\}_{i=1}^{N}$. Trivial regions occupying less than 1\% of the image are discarded, typically resulting in 2–5 valid masks per image.

Both the generated regions and atomic sentences are encoded using the pre-trained CLIP (ViT-B/16) encoders~\cite{radford2021learning} to extract their respective features:
\begin{equation}
\boldsymbol{i}_l^{(i)} = \mathrm{f}^{\mathcal{I}}(\mathcal{I}_l^{(i)}), \quad 
\boldsymbol{t}_l^{(j)} = \mathrm{f}^{\mathcal{T}}(\mathcal{T}_l^{(j)}),
\end{equation}
We then compute the cosine similarity between all region–sentence combinations and select the pair with the highest score:
\begin{equation}
(i^*, j^*) = \arg\max_{i,j} \, \cos\!\left( \boldsymbol{i}_{\mathrm{l}}^{(i)}, \boldsymbol{t}_{\mathrm{l}}^{(j)} \right).
\label{eq:optimal_pair}
\end{equation}
This process uses a greedy assignment strategy to ensure a one-to-one alignment between regions and sentences. If the selected region $\mathcal{I}_l^{(i^*)}$ covers nearly the entire image, it is discarded to maintain locality. Otherwise, the aligned pseudo pair $(\mathcal{I}_l, \mathcal{T}_l) = (\mathcal{I}_l^{(i^*)}, \mathcal{T}_l^{(j^*)})$ is used for fine-grained supervision. This automatic procedure allows the model to leverage localized visual cues without relying on labor-intensive manual annotations.

\subsubsection{P-VLG Benchmark} 
\label{subsec:pvlg}  

\textbf{Construction and Data Provenance.} To facilitate large-scale fine-grained learning while adhering to ethical data standards, we introduce the \textbf{P-VLG Benchmark}. The images in this benchmark are curated from four established public datasets: CUHK-PEDES~\cite{li2017person}, ICFG-PEDES~\cite{ding2021semantically}, RSTPReid~\cite{zhu2021dssl}, and UfineBench~\cite{zuo2024ufinebench}. By leveraging these existing resources, we ensure compliance with privacy protocols while introducing rich, multi-granular annotations that overcome the semantic sparsity of original short captions.

The benchmark contains 6,801 training identities with 48,485 images and 68,990 enriched long-form captions. The validation set includes over 100,000 region–caption alignments, providing a valuable resource for evaluating local grounding. The test set consists of 400 unseen identities with 6,658 image–caption pairs, supporting both global and region-level retrieval evaluation.

\textbf{Annotation Process.} All region–caption pairs are automatically generated using the \textit{Region-to-Sentence Matching (RSM)} module. In this process, long-form captions are decomposed into atomic sentences, while images are segmented into semantic regions using SAM~\cite{SAM}. CLIP~\cite{clip} features are extracted for both sentences and regions, and the most similar sentence-region pairs are selected to form pseudo region–sentence alignments. This automated approach provides dense, high-quality supervision without manual annotation, making P-VLG the first TBPS benchmark to support a multi-granular evaluation protocol that jointly assesses global and local alignment performance.

\subsection{The ROGLE Framework}
\label{subsec:framework}

ROGLE synergizes coarse and fine-grained learning via three components: a dual-encoder backbone, a reliability-guided mechanism for noise robustness, and a collaborative objective for multi-granular alignment.

\subsubsection{Dual-Encoder Feature Extraction}
\label{subsubsec:feature_extraction}

Our model adopts a dual-encoder architecture initialized with pre-trained CLIP weights~\cite{radford2021learning}, leveraging CLIP’s strong cross-modal alignment capabilities. The architecture consists of two encoders: an image encoder \( f^v(\cdot) \) and a text encoder \( f^t(\cdot) \).

\paragraph{Image Encoder}  
Given an input image \( \mathcal{I} \in \mathbb{R}^{H \times W \times C} \), the image encoder—implemented as a Vision Transformer (ViT)—first partitions the image into a sequence of non-overlapping patches. Each patch is linearly embedded into a token, and a learnable \texttt{[CLS]} token is prepended to the sequence. The resulting token sequence is then processed through a stack of Transformer layers. The final representation of the \texttt{[CLS]} token is projected into the joint embedding space to yield the global image embedding:
\begin{equation}
\mathbf{f}^v = f^v(\mathcal{I}).
\end{equation}

\paragraph{Text Encoder}  
For an input text description \( \mathcal{T} \), the text encoder—a Transformer-based model—first tokenizes the text using Byte Pair Encoding (BPE), as in CLIP. Special tokens \texttt{[SOS]} (start-of-sequence) and \texttt{[EOS]} (end-of-sequence) are added to frame the token sequence. The encoder processes this sequence through its Transformer layers, and the final hidden state at the \texttt{[EOS]} position is projected into the same joint embedding space to produce the global text embedding:
\begin{equation}
    \mathbf{f}^t = f^t(\mathcal{T}).
\end{equation}

\subsubsection{Reliability-Guided Alignment }
\label{subsubsec:reliability}

Real-world datasets often contain noisy correspondences. To address this, we adopt a reliability-guided calibration mechanism inspired by RDE~\cite{qin2024noisy} to dynamically handle uncertain training pairs.

For each training pair $(\mathcal{I}_i, \mathcal{T}_i)$ in a batch of size $N_b$, we compute its alignment loss, $\ell_i = \mathcal{L}(\mathcal{I}_i, \mathcal{T}_i)$. We then fit a two-component Gaussian Mixture Model (GMM) to the loss distributions of both long and short captions to estimate the probability that a pair is reliable ($k=1$) or unreliable ($k=0$). Based on a consensus between these two estimates, we partition the data into three sets: clean ($\hat{\mathcal{P}}^c$), noisy ($\hat{\mathcal{P}}^n$), and uncertain ($\hat{\mathcal{P}}^u$).

Finally, a reliability-aware weight $\hat{w}_{i}$ is assigned to each sample:
\begin{equation}
\hat{w}_{i} =
\begin{cases}
2, & \text{if } (\mathcal{I}_i, \mathcal{T}_i) \in \hat{\mathcal{P}}^c, \\
0, & \text{if } (\mathcal{I}_i, \mathcal{T}_i) \in \hat{\mathcal{P}}^n, \\
1, & \text{otherwise}.
\end{cases}
\end{equation}
This allows the model to prioritize learning from clean pairs while reducing the influence of noisy ones.

\subsubsection{Global-Local Collaborative Learning}
\label{subsubsec:collab}

ROGLE learns features at multiple granularities by jointly supervising coarse and fine-grained alignment. Given global pairs $(\mathcal{I}_g, \mathcal{T}_g)$ and local pseudo-pairs $(\mathcal{I}_l, \mathcal{T}_l)$ from our RSM module, the encoder extracts their respective embeddings:
\begin{align}
\mathbf{f}^v_g &= f^v(\mathcal{I}_g), \quad \mathbf{f}^t_g = f^t(\mathcal{T}_g) \\
\mathbf{f}^v_l &= f^v(\mathcal{I}_l), \quad \mathbf{f}^t_l = f^t(\mathcal{T}_l)
\end{align}

The learning is driven by a dual-branch loss:
    \textbf{Global Loss ($\mathcal{L}_{\text{global}}$):} This loss facilitates coarse-grained alignment between the global embeddings of entire images and captions. Following IRRA~\cite{IRRA}, it is defined as:
\begin{equation}
    \mathcal{L}_{\text{global}} = \mathcal{L}_{\text{SDM}} + \mathcal{L}_{\text{InfoNCE}}
\end{equation}
\textbf{Local Loss ($\mathcal{L}_{\text{local}}$):} This loss ensures fine-grained alignment between local image regions and sentence fragments. It is a contrastive loss over the local embeddings in a batch:
{\small
\begin{equation}
\mathcal{L}_{\text{local}} =
- \frac{1}{N_b} \sum_{n=1}^{N_b}
\log \frac{
\exp\!\big(\cos(\mathbf{f}^{v,(n)}_l, \mathbf{f}^{t,(n)}_l) / \tau\big)
}{
\sum_{k=1}^{N_b} \exp\!\big(\cos(\mathbf{f}^{v,(n)}_l, \mathbf{f}^{t,(k)}_l) / \tau\big)
}.
\label{eq:local_loss}
\end{equation}
}

    where $\cos(\cdot, \cdot)$ is the cosine similarity and $\tau$ is a temperature parameter.

\subsubsection{Training Objective}
\label{subsubsec:training}

The overall training objective combines the global and local losses, weighted by the reliability score, to create a robust, multi-granular learning framework. The total loss is:
\begin{equation}
\mathcal{L}_{\text{total}} = \frac{1}{N_b} \sum_{i=1}^{N_b} \hat{w}_{i} \mathcal{L}_{\text{global}}^{(i)} + \mathcal{L}_{\text{local}}
\end{equation}
Here, $\mathcal{L}_{\text{global}}^{(i)}$ is the global loss for the $i$-th sample. This two-branch supervision enables ROGLE to capture both global semantic coherence and fine-grained regional details, resulting in precise and robust text-to-image person retrieval.

% 表格
% ***********************************

% ***********************************
% **********************************

% 跨数据集
% ***********************************

\begin{table*}[t]
% \caption{Performance Evaluation on Image-Text Retrieval. Bold: best results, Underline: second best results.}
\caption{Performance comparisons on the CUHK-PEDES dataset. The best results are in \textbf{bold}.} 
  \label{tab:1}
\begin{center}
\resizebox{1.01\textwidth}{0.25\textheight}{ 
\begin{tabular}{l|lcc|ccccc}\toprule[1pt] 
Methods            & Ref.      & Image Enc. & Text Enc.    & R-1 & R-5 & R-10 & mAP   & mINP  \\\midrule
% CMPM/C~\cite{zhang2018deep}& ECCV'18   & RN50       & LSTM         & 49.37  & -      & 79.27   & -     & -     \\
% TIMAM~\cite{sarafianos2019adversarial}              & ICCV'19   & RN101      & BERT         & 54.51  & 77.56  & 79.27   & -     & -     \\
 ViTAA~\cite{wang2020vitaa}              & ECCV'20   & RN50       & LSTM         & 54.92  & 75.18  & 82.90   & 51.60 & -     \\

 DSSL~\cite{zhu2021dssl}               & ACMMM'21     & RN50       & BERT         & 59.98  & 80.41  & 87.56   & -     & -     \\
SSAN~\cite{ding2021semantically}               & arXiv'21  & RN50       & LSTM         & 61.37  & 80.15  & 86.73   &       & -     \\
Lapscore~\cite{wu2021lapscore}           & ICCV'21   & RN50       & BERT         & 63.40   & -      & 87.80   & -     & -     \\
% ISANet~\cite{yan2022image}             & arXiv'22  & RN50       & LSTM         & 63.92  & 82.15  & 87.69   & -     & -     \\

% Han et al. ~\cite{han2021text} & BMVC'21   & CLIP-RN101 & CLIP-Xformer & 64.08  & 81.73  & 88.19   & 60.08 & -     \\
% SAF~\cite{li2022learning}                & ICASSP'22 & ViT-Base   & BERT         & 64.13  & 82.62  & 88.40   & -     & -     \\
LBUL~\cite{wang2022look}               & ACMMM'22     & RN50       & BERT         & 64.04  & 82.66  & 87.22   & -     & -     \\
% CAIBC~\cite{wang2022caibc}              & ACMMM'22     & RN50       & BERT         & 64.43  & 82.87  & 88.37   & -     & -     \\
% AXM-Net~\cite{farooq2022axm}            & ACMMM'22     & RN50       & BERT         & 64.44  & 80.52  & 86.77   & 58.73 & -     \\
LGUR~\cite{shao2022learning}               & ACMMM'22     & DeiT-Small & BERT         & 65.25  & 83.12  & 89.00   & -     & -     \\
DCEL~\cite{qin2022deepdecl}&ACMMM'22&CLIP-ViT& CLIP-Xformer &   71.36&  {88.11}&  {92.48}& 64.25& 48.26\\
IVT~\cite{shu2022see}                & ECCV'22  & ViT-Base   & BERT         & 65.59  & 83.11  & 89.21   & -     & -     \\
CFine~\cite{yan2023clip}              & TIP'23  & CLIP-ViT   & BERT         & 69.57  & 85.93  & 91.15   & -     & -     \\
IRRA~\cite{jiang2023cross}               & CVPR'23   & CLIP-ViT   & CLIP-Xformer & 73.38  & 89.93  & 93.71   & 66.13 & 50.24 \\
% \textbf{Our RDE-BGE}&-&CLIP-ViT&CLIP-Xformer&74.20&89.57&93.49&66.12&49.62 \\
% \textbf{Our RDE-TGE}&-&CLIP-ViT&CLIP-Xformer&74.85&89.80&93.96&67.23&51.25 \\

 BiLMa~\cite{fujii2023bilma}    & ICCV'23   & CLIP-ViT   & CLIP-Xformer  &74.03& 89.59& 93.62 &66.57&-\\ 

PBSL~\cite{shen2023pedestrian}&ACMMM'23 & RN50&BERT&65.32 &83.81 &89.26&-&-\\
BEAT\cite{ma2023beat}&ACMMM'23&RN101&BERT& 65.61& 83.45&89.54&-&-\\ 
LCR$^2$S~\cite{yan2023learning}&ACMMM'23&RN50&TextCNN&67.36& 84.19& 89.62& 59.24&\\
DCEL~\cite{li2023dcel}&ACMMM'23&CLIP-ViT& CLIP-Xformer &   75.02&  {90.89}&  {94.52}& -& -\\
APTM~\cite{yang2023towards}&ACMMM'23&Swin-Transformer& Bert &76.53	&90.04	&94.15	&66.91\\
UniPT~\cite{shao2023unified}&ICCV'23&CLIP-ViT& CLIP-Xformer &68.50& 84.67 &-&-&-\\ 

 RaSa~\cite{bai2023rasa}& IJCAI'23& ALBEFF& ALBEFF&76.51& 90.29 & 94.25 &69.38&\\

TBPS~\cite{cao2024empirical}&AAAI'24&CLIP-ViT& CLIP-Xformer&73.54& 88.19 &92.35& 65.38&-\\
DP~\cite{song2024diverse} & AAAI'24 & CLIP-ViT & CLIP-Xformer & 75.66 & 90.59 & 94.07 & 66.58 & - \\

IRRA+IRLT~\cite{liu2024causality} & AAAI'24 & CLIP-ViT & CLIP-Xformer & 74.46 & 90.19 & 94.01 & - & - \\
UASA~\cite{zhao2024unifying} & AAAI'24 & CLIP-ViT & CLIP-Xformer & 74.25 & 89.83 &  93.58 & 66.15 & - \\

{RDE}~\cite{qin2024noisy} & CVPR'24 & CLIP-ViT& CLIP-Xformer & {75.94} &{90.14}& {94.12} &{67.56}& {51.44} \\  

CFAM ~\cite{zuo2024ufinebench}&CVPR'24&CLIP-ViT& CLIP-Xformer&75.60& 90.53&-&67.27&-\\

PLOT ~\cite{park2024plottextbasedpersonsearch}&ECCV'24&CLIP-ViT& CLIP-Xformer&75.28&  90.42& 94.12 &-& - \\

% MLLM+IRRA ~\cite{tan2024harnessingmllm}&CVPR'24&CLIP-ViT& CLIP-Xformer&76.82&  91.16&-&69.55&-\\
MGRL ~\cite{lv2024mgrl}&ICASSP'24&CLIP-ViT& CLIP-Xformer&73.91& 90.68&-& 67.28&-\\
DM-Adapter~\cite{liu2025dm} & AAAI'25 & CLIP-ViT & CLIP-Xformer & 72.17 & 88.74 & 92.85 &64.33 & - \\
OCDL ~\cite{li2025object}&ICASSP'25&CLIP-ViT& CLIP-Xformer&75.10& 89.43&-& 68.18&-\\

\midrule
\textbf{ ROGLE (Ours) }& -& CLIP-ViT& CLIP-Xformer & \textbf{ 76.93} & \textbf{ 91.24}&\textbf {95.00} &\textbf{69.14}& \textbf{53.23} \\  

\bottomrule[1pt]
\end{tabular}}
% \caption{Performance comparisons on the CUHK-PEDES dataset. The best results are in \textbf{bold}.} 

\end{center}
\end{table*}

\section{Experiments}
In this section, we present extensive experiments to demonstrate the effectiveness and advantages of the proposed ROGLE framework.

\subsection{Datasets and Settings}
\textbf{CUHK-PEDES} \cite{li2017person} is the first dataset for text-to-image person retrieval, containing 40,206 images and 80,412 descriptions for 13,003 identities, with 11,003 identities for training and 1,000 identities each for validation and testing. 
\textbf{ICFG-PEDES} \cite{ding2021semantically} includes 54,522 image–text pairs for 4,102 identities, where 3,102 identities are used for training and 1,000 for testing. 
\textbf{RSTPReid} \cite{zhu2021dssl} consists of 20,505 images of 4,101 identities captured by 15 cameras, each image paired with two textual descriptions, and is split into 3,701/200/200 identities for training, validation, and testing, respectively.  

\subsection{Implementation Details}

ROGLE is built on CLIP~\cite{radford2021learning} using a ViT-B/16 image encoder and a Transformer text encoder, both initialized from the official 400M-pretrained weights.Region are generated by SAM~\cite{kirillov2023segment} (ViT-H) in automatic mode, producing 2–5 valid masks per image after filtering trivial regions.Captions are decomposed into short atomic sentences using spaCy~\cite{honnibal2017spacy}, and region–sentence matching is computed via cosine similarity with greedy assignment to ensure one-to-one alignment.
Images are augmented by random flip, crop, erase, and color jitter, then resized to $384 \times 128$.
Texts are tokenized by CLIP’s BPE (max length 77).
The model is trained for 60 epochs with Adam~\cite{kingma2015adam} (batch size 64, weight decay $5!\times!10^{-4}$).
A two-stage learning rate schedule is used: linear warm-up for 5 epochs, then cosine decay.
The contrastive temperature $\tau$ is 0.02.
The reliability calibration updates a GMM every 5 epochs (from epoch 10) with threshold 0.5.
All experiments run on a single RTX 4090 GPU.
Training takes 8–10 hours per dataset, and inference speed reaches 128 image–text pairs per second.The model has 158M parameters.

\subsection{Evaluation Protocols}
We adopt the standard Rank-K metrics (K = 1, 5, 10) to evaluate retrieval performance. In addition, following \cite{jiang2023cross}, we include mean Average Precision (mAP) and mean Inverse Negative Penalty (mINP) to offer a more complete evaluation of model effectiveness.

\subsection{Comparison with State-of-the-Art}

\textbf{Performance on CUHK-PEDES.} 
We first evaluate ROGLE on the widely used CUHK-PEDES. As reported in \cref{tab:1}, ROGLE achieves 76.93\% Rank-1 accuracy, outperforming all existing state-of-the-art methods. This performance gain highlights the effectiveness of incorporating both global and local supervision during training.

\begin{table}[h]
\centering

\caption{Performance comparisons on the ICFG-PEDES dataset. The best results are in \textbf{bold}. }
\resizebox{1.03 \linewidth}{!}{

\begin{tabular}{l|ccccc}
\toprule[1pt] 
Methods    & R-1 & R-5 & R-10 & mAP   & mINP \\\midrule
Dual Path~\cite{zheng2020dual} & 38.99  & 59.44  & 68.41   & -     & -    \\
CMPM/C~\cite{zhang2018deep}    & 43.51  & 65.44  & 74.26   & -     & -    \\
ViTAA~\cite{wang2020vitaa}     & 50.98  & 68.79  & 75.78   & -     & -    \\
SSAN~\cite{ding2021semantically}      & 54.23  & 72.63  & 79.53   & -     & -    \\
IVT~\cite{shu2022see}       & 56.04  & 73.60  & 80.22   & -     & -    \\
ISANet~\cite{yan2022image}    & 57.73  & 75.42  & 81.72   & -     & -    \\
CFine~\cite{yan2023clip}     & 60.83  & 76.55  & 82.42   & -     & -    \\
IRRA~\cite{jiang2023cross}      & 63.46  & 80.25  & 85.82   & 38.06 & {7.93} \\
BiLMa~\cite{fujii2023bilma}  &
63.83& 80.15& 85.74& 38.26&-\\
PBSL~\cite{shen2023pedestrian}&57.84 &75.46& 82.15&-&-\\
BEAT\cite{ma2023beat}&58.25& 75.92 &81.96&-&-\\
LCR$^2$S~\cite{yan2023learning}&57.93 &76.08& 82.40 &38.21&-\\
DCEL~\cite{li2023dcel}&64.88& 81.34 &86.72&-&-\\
UniPT~\cite{shao2023unified}&60.09 &76.19&-&-&-\\
    
RaSa~\cite{bai2023rasa}&   65.28&    80.40 &   85.12 &   41.29&-\\
TBPS~\cite{cao2024empirical}&65.05& 80.34 &85.47& 39.83&-\\
CFAM ~\cite{zuo2024ufinebench}&65.38& 81.17&-&39.42&-\\

PLOT ~\cite{park2024plottextbasedpersonsearch} &65.76  &81.39  &86.73 &-& - \\

OCDL ~\cite{li2025object}&64.53& 80.23&-& 40.76&-\\

\midrule

\textbf{ROGLE(Ours)} &\textbf{66.97}& \textbf{82.27}& \textbf{87.21}& \textbf{39.81} &\textbf{7.88} \\
\bottomrule[1pt]
\end{tabular}}
  \label{tab:icfg}
\end{table}

\textbf{Performance on ICFG-PEDES.} 
We further evaluate our method on the ICFG-PEDES. As shown in \cref{tab:icfg}, ROGLE achieves 66.97\% Rank-1 accuracy, surpassing prior methods such as IRRA (63.46\%) by a significant margin. In addition to Rank-1 accuracy, our model also shows improvements in mAP and mINP, suggesting that ROGLE is better at retrieving harder samples and handling more challenging matching scenarios.

\begin{table}[h]
\centering 

\caption{\textbf{Performance Comparison on the RSTPReid Dataset.} The best results are highlighted in \textbf{bold}, while the second-best results are underlined.}
\resizebox{1.03 \linewidth}{!}{
\begin{tabular}{l|ccccc}
\toprule[1pt] 
Methods & R-1& R-5 & R-10& mAP   & mINP  \\\midrule
DSSL~\cite{zhu2021dssl}   & 39.05& 62.60  & 73.95& -     & -     \\
SSAN~\cite{ding2021semantically}   & 43.50& 67.80  & 77.15& -     & -     \\
LBUL~\cite{wang2022look}   & 45.55& 68.20  & 77.85& -     & -     \\
IVT~\cite{shu2022see}    & 46.70& 70.00  & 78.80& -     & -     \\
CFine~\cite{yan2023clip}  & 50.55& 72.50  & 81.60& -     & -     \\
IRRA~\cite{jiang2023cross}  & 60.20 & 81.30  & 88.20 & 47.17 & 25.28 \\ 
BiLMA~\cite{fujii2023bilma}&61.20& 81.50& 88.80 &48.51&-\\
PBSL~\cite{shen2023pedestrian}&47.80& 71.40& 79.90&-&-\\
BEAT\cite{ma2023beat}&48.10 &73.10 &81.30&-&-\\
LCR$^2$S~\cite{yan2023learning}&54.95& 76.65& 84.70 &40.92&-\\
DCEL~\cite{li2023dcel}& 61.35& 83.95 &{90.45}&-&-\\
    RaSa~\cite{bai2023rasa}&  \textbf{ 66.90} &   \textbf{ 86.50 } &  \textbf{  91.35 }&  \textbf{  52.31}&-\\
TBPS~\cite{cao2024empirical} &61.95& 83.55& 88.75& 48.26&-\\
CFAM ~\cite{zuo2024ufinebench}&62.45& 83.55&-&49.50&-\\

PLOT ~\cite{park2024plottextbasedpersonsearch} &61.80 &82.85 &89.45 &-& - \\
OCDL ~\cite{li2025object}&61.60&82.35&-& 49.77&-\\

\midrule
\textbf{ROGLE (Ours)} &\underline{65.20}& \underline{83.75} &\underline{ 89.45}& \underline{50.93}& \underline{27.83 } \\
\bottomrule[1pt]
\end{tabular}}
  \label{tab:rstp}
\end{table}

\textbf{Performance on RSTPReid.}
We conduct experiments on the RSTPReid. As shown in \cref{tab:rstp}, ROGLE achieves 65.20\% Rank-1 accuracy, significantly higher than previous methods such as IVT (51.7\%) and Cfine (55.55\%). This demonstrates the superior ability of our approach to handle high-resolution, complex textual descriptions in crowded urban environments. The integration of sentence-level supervision and region-level alignment helps capture subtle semantic cues, crucial for this dataset.

\begin{table}[h]
\centering
\caption{Performance comparisons on the P-VLG dataset. The best results are in \textbf{bold}. }
\resizebox{1 \linewidth}{!}{

\begin{tabular}{l|ccccc}
\toprule[1pt] 
Methods    & R-1 & R-5 & R-10 & mAP   & mINP \\\midrule
DECL~\cite{qin2022deep}      & 73.78  & 86.66  & 90.80  & 53.09     & 19.72    \\

IRRA~\cite{jiang2023cross}      & 73.30  & 86.57 & 90.95   & 54.12& 21.22 \\

RDE~\cite{qin2024noisy}      & 68.82  &83.66 & 88.72   & 48.10& 15.75 \\
  
\midrule

\textbf{ROGLE(Ours)} &\textbf{78.03}& \textbf{89.17}& \textbf{92.60}& \textbf{57.82} &\textbf{ 24.01} \\

\bottomrule[1pt]
\end{tabular}}
\label{tab:vlg}
\end{table}

\subsection{Performance on Long Captions}

To validate ROGLE's effectiveness on detailed, multi-attribute queries, we evaluate on the P-VLG benchmark, which exclusively features long-form captions with an average length of 62.3 words—significantly longer than traditional TBPS datasets such as CUHK-PEDES. As shown in \cref{tab:vlg}, ROGLE achieves 78.03\% Rank-1 accuracy, establishing a new state-of-the-art on this challenging long-caption benchmark. Compared with strong baselines IRRA (73.30\%) and RDE (68.82\%), our method delivers substantial improvements of +4.73\% and +9.21\%, respectively. These significant gains demonstrate that ROGLE's region-sentence alignment mechanism is particularly effective for handling complex, compositional descriptions that require fine-grained visual grounding. The superior performance can be attributed to two key factors: (1) RSM automatically decomposes long captions into atomic sentences, each aligned with specific image regions, enabling fine-grained correspondences that would be obscured in holistic global matching; (2) the dual-branch multi-granular learning jointly optimizes global identity-level coherence and local attribute-level details, critical when queries contain multiple compositional attributes. In contrast, baseline methods relying primarily on global [CLS] token alignment struggle to capture the nuanced semantic details present in long-form descriptions. These results validate ROGLE's robustness and suitability for real-world person search scenarios, where witness reports often contain detailed, compositional information spanning multiple visual attributes and contextual details.

\subsection{Ablation Study}

We validate ROGLE on CUHK-PEDES (Table \ref{tab:ablation}). The global-only baseline achieves 71.56\% Rank-1, while local-only alignment yields 69.92\%, indicating neither is sufficient alone. Combining both strategies significantly improves accuracy to 74.87\% (+3.31\%), demonstrating their complementary nature. Furthermore, adding the RGAP module boosts Rank-1 to 76.93\% (+2.06\%) by adaptively filtering noisy pseudo-labels via Beta Mixture Modeling. These results confirm that effective TBPS requires synergizing global coherence, local discriminative details, and noise-robust training.

\begin{table}[htbp]
\centering
\caption{Ablation study of ROGLE components on CUHK-PEDES. The best results are in \textbf{bold}.}
\label{tab:ablation}
\resizebox{ 1.05 \linewidth}{!}{
\begin{tabular}{l|ccc|ccc}
\toprule
\multirow{2}{*}{Method} & \multicolumn{3}{c|}{Components}  & \multicolumn{3}{c}{Performance} \\
\cline{2-4} \cline{5-7}
 & $\mathcal{L}_{\text{global}}$ & $\mathcal{L}_{\text{local}}$ & RGAP & R-1 & R-5 & mAP \\
\midrule
Global only             & \checkmark &            &       & 71.56 & 88.23 & 64.38 \\
Local only              &            & \checkmark &       & 69.92 & 87.15 & 63.12 \\
Global + Local          & \checkmark & \checkmark &       & 74.87 & 90.02 & 67.25 \\
ROGLE (full)            & \checkmark & \checkmark & \checkmark & \textbf{76.93} & \textbf{91.24} & \textbf{69.14} \\
\bottomrule
\end{tabular}
}
\end{table}

\subsection{Validation of Pseudo-Label Quality}
\label{subsec:rsm_validation}

To address concerns regarding the reliability of the automated Region-to-Sentence Matching (RSM) module, we conducted a human verification study. Specifically, we randomly sampled 200 generated region-sentence pairs from the P-VLG training set. Human annotators were tasked with a binary judgment: determining whether the decomposed textual sentence accurately described the visual content within the corresponding image region cropped by SAM.
The evaluation results indicated a matching accuracy of \textbf{87.5\%}. This empirical evidence demonstrates that despite the potential global representational bias of the pre-trained CLIP model, our RSM strategy—aided by SAM's precise semantic segmentation—is capable of mining high-quality fine-grained supervision signals. These reliable pseudo-labels are instrumental in the model's superior performance on long-form query retrieval.
\section{Conclusion}
 In this paper, we propose \textbf{ROGLE}, a unified framework that advances fine-grained text-based person search via automatic region--sentence alignment and multi-granular learning. To alleviate the lack of fine-grained supervision, we introduce the RSM module, which automatically mines reliable region--sentence pseudo-pairs from long-form captions without manual annotation; building on this module, we design a multi-granularity supervision strategy that combines global contrastive learning with region-level alignment. We also introduce the \textbf{P-VLG Benchmark}, a large-scale dataset with enriched long-form captions and more than 100{,}000 region-level annotations; as the first TBPS benchmark that supports both global and local evaluation protocols, P-VLG enables comprehensive and rigorous assessment of fine-grained alignment. Extensive experiments show that ROGLE achieves state-of-the-art performance on multiple challenging benchmarks, including CUHK-PEDES, ICFG-PEDES, RSTPReid, and P-VLG, and remains robust under cross-domain shifts, validating the effectiveness and scalability of our approach and positioning ROGLE as a practical solution for reliable fine-grained person search in real-world applications.

\section{Limitations}
\label{sec:limitations}

While ROGLE and the P-VLG benchmark advance TBPS, limitations remain. First, the greedy one-to-one assignment strategy in RSM may not optimally handle complex descriptions where a single sentence refers to multiple regions. Second, our evaluation is currently limited to English-language data. Adapting ROGLE to multilingual settings remains an important direction for future work.

\section*{Ethics Statement}
\label{sec:ethics}

We prioritize ethical considerations and data privacy in the construction of the P-VLG benchmark. We explicitly clarify that this work \textbf{does not involve the collection of new visual data or surveillance footage}. All pedestrian images in the P-VLG benchmark are curated exclusively from four established, publicly available datasets: CUHK-PEDES~\cite{li2017person}, ICFG-PEDES~\cite{ding2021semantically}, RSTPReid~\cite{zhu2021dssl}, and UfineBench~\cite{zuo2024ufinebench}.

We strictly adhere to the original data usage policies, license agreements, and privacy protocols of these source datasets. No new personally identifiable information (PII) was collected, and no human subjects were recruited for visual recording. Our contribution focuses solely on generating enriched textual annotations and region-level alignments for these existing public images to advance fine-grained research.
% \section{Acknowledgment}

% \section*{Acknowledgments}
% \begin{sloppypar}
% This work was supported by the ``Pioneer'' and ``Leading Goose'' R\&D Program of Zhejiang under (Grant No. 2025C02110), Public Welfare Research Program of Ningbo under (Grant No. 2024S062), and Yongjiang Talent Project of Ningbo under (Grant No. 2024A-161-G).    
% \end{sloppypar}

\bibliography{main}

\begin{thebibliography}{53}
\providecommand{\natexlab}[1]{#1}

\bibitem[{Bai et~al.(2023)Bai, Cao, Gao, Cao, Chen, Fan, Nie, and Zhang}]{bai2023rasa}
Yang Bai, Min Cao, Daming Gao, Ziqiang Cao, Chen Chen, Zhenfeng Fan, Liqiang Nie, and Min Zhang. 2023.
\newblock Rasa: relation and sensitivity aware representation learning for text-based person search.
\newblock In \emph{Proceedings of the Thirty-Second International Joint Conference on Artificial Intelligence}, pages 555--563.

\bibitem[{Cao et~al.(2024)Cao, Bai, Zeng, Ye, and Zhang}]{cao2024empirical}
Min Cao, Yang Bai, Ziyin Zeng, Mang Ye, and Min Zhang. 2024.
\newblock An empirical study of clip for text-based person search.

\bibitem[{Chen et~al.(2024)Chen, Lai, Zhang, Wang, Eichner, You, Cao, Zhang, Yang, and Gan}]{chen2024contrastive}
Hong-You Chen, Zhengfeng Lai, Haotian Zhang, Xinze Wang, Marcin Eichner, Keen You, Meng Cao, Bowen Zhang, Yinfei Yang, and Zhe Gan. 2024.
\newblock Contrastive localized language-image pre-training.
\newblock \emph{arXiv preprint arXiv:2410.02746}.

\bibitem[{Ding et~al.(2021)Ding, Ding, Shao, and Tao}]{ding2021semantically}
Zefeng Ding, Changxing Ding, Zhiyin Shao, and Dacheng Tao. 2021.
\newblock Semantically self-aligned network for text-to-image part-aware person re-identification.
\newblock \emph{arXiv preprint arXiv:2107.12666}.

\bibitem[{Feng et~al.(2026)Feng, Cai, Xie, Wu, and Jin}]{Feng_Cai_Xie_Wu_Jin_2026}
Fangming Feng, Sihang Cai, Zequn Xie, Yangyang Wu, and Tao Jin. 2026.
\newblock \href {https://doi.org/10.1609/aaai.v40i5.37392} {Scene-aware spatiotemporal generalization: Towards robust temporal action detection across domains}.
\newblock \emph{Proceedings of the AAAI Conference on Artificial Intelligence}, 40(5):3903–3911.

\bibitem[{Fujii and Tarashima(2023)}]{fujii2023bilma}
Takuro Fujii and Shuhei Tarashima. 2023.
\newblock Bilma: Bidirectional local-matching for text-based person re-identification.
\newblock In \emph{Proceedings of the IEEE/CVF International Conference on Computer Vision}, pages 2786--2790.

\bibitem[{Gao et~al.(2021)Gao, Cai, Jiang, Zheng, Zhang, Gong, Peng, Guo, and Sun}]{gao2021contextual}
Chenyang Gao, Guanyu Cai, Xinyang Jiang, Feng Zheng, Jun Zhang, Yifei Gong, Pai Peng, Xiaowei Guo, and Xing Sun. 2021.
\newblock Contextual non-local alignment over full-scale representation for text-based person search.
\newblock \emph{arXiv preprint arXiv:2101.03036}.

\bibitem[{He et~al.(2021)He, Luo, Wang, Wang, Li, and Jiang}]{he2021transreid}
Shuting He, Hao Luo, Pichao Wang, Fan Wang, Hao Li, and Wei Jiang. 2021.
\newblock Transreid: Transformer-based object re-identification.
\newblock In \emph{ICCV}, pages 15013--15022.

\bibitem[{Honnibal and Montani(2017)}]{honnibal2017spacy}
Matthew Honnibal and Ines Montani. 2017.
\newblock spacy 2: Natural language understanding with bloom embeddings, convolutional neural networks and incremental parsing.
\newblock \emph{To appear}, 7(1):411--420.

\bibitem[{Jiang and Ye(2023{\natexlab{a}})}]{IRRA}
D.~Jiang and M.~Ye. 2023{\natexlab{a}}.
\newblock Cross-modal implicit relation reasoning and aligning for text-to-image person retrieval.
\newblock In \emph{Proceedings of the IEEE/CVF Conference on Computer Vision and Pattern Recognition}, pages 2787--2797.

\bibitem[{Jiang and Ye(2023{\natexlab{b}})}]{jiang2023cross}
Ding Jiang and Mang Ye. 2023{\natexlab{b}}.
\newblock Cross-modal implicit relation reasoning and aligning for text-to-image person retrieval.
\newblock In \emph{CVPR}, pages 2787--2797.

\bibitem[{Kingma and Ba(2015)}]{kingma2015adam}
Diederik~P Kingma and Jimmy Ba. 2015.
\newblock Adam: A method for stochastic optimization.
\newblock In \emph{ICLR (Poster)}.

\bibitem[{Kirillov et~al.(2023{\natexlab{a}})Kirillov, Mintun, Ravi, Mao, Rolland, Gustafson, Xiao, Whitehead, Berg, Lo, Doll{\'a}r, and Girshick}]{SAM}
Alexander Kirillov, Eric Mintun, Nikhila Ravi, Hanzi Mao, Chloe Rolland, Laura Gustafson, Tete Xiao, Spencer Whitehead, Alexander~C. Berg, Wan-Yen Lo, Piotr Doll{\'a}r, and Ross Girshick. 2023{\natexlab{a}}.
\newblock Segment anything.
\newblock \emph{arXiv:2304.02643}.

\bibitem[{Kirillov et~al.(2023{\natexlab{b}})Kirillov, Mintun, Ravi, Mao, Rolland, Gustafson, Xiao, Whitehead, Berg, Lo et~al.}]{kirillov2023segment}
Alexander Kirillov, Eric Mintun, Nikhila Ravi, Hanzi Mao, Chloe Rolland, Laura Gustafson, Tete Xiao, Spencer Whitehead, Alexander~C Berg, Wan-Yen Lo, et~al. 2023{\natexlab{b}}.
\newblock Segment anything.
\newblock In \emph{Proceedings of the IEEE/CVF international conference on computer vision}, pages 4015--4026.

\bibitem[{Li et~al.(2025)Li, Liu, Su, and Zhao}]{li2025object}
Haiwen Li, Delong Liu, Fei Su, and Zhicheng Zhao. 2025.
\newblock Object-centric discriminative learning for text-based person retrieval.
\newblock In \emph{ICASSP 2025-2025 IEEE International Conference on Acoustics, Speech and Signal Processing (ICASSP)}, pages 1--5. IEEE.

\bibitem[{Li et~al.(2023)Li, Xu, Yang, Shen, Mo, Li, and Shen}]{li2023dcel}
Shenshen Li, Xing Xu, Yang Yang, Fumin Shen, Yijun Mo, Yujie Li, and Heng~Tao Shen. 2023.
\newblock Dcel: Deep cross-modal evidential learning for text-based person retrieval.
\newblock In \emph{Proceedings of the 31st ACM International Conference on Multimedia}, pages 6292--6300.

\bibitem[{Li et~al.(2017)Li, Xiao, Li, Zhou, Yue, and Wang}]{li2017person}
Shuang Li, Tong Xiao, Hongsheng Li, Bolei Zhou, Dayu Yue, and Xiaogang Wang. 2017.
\newblock Person search with natural language description.
\newblock In \emph{CVPR}, pages 1970--1979.

\bibitem[{Liu et~al.(2025)Liu, Liu, Lan, Yang, Li, and Liao}]{liu2025dm}
Yating Liu, Zimo Liu, Xiangyuan Lan, Wenming Yang, Yaowei Li, and Qingmin Liao. 2025.
\newblock Dm-adapter: Domain-aware mixture-of-adapters for text-based person retrieval.
\newblock In \emph{Proceedings of the AAAI Conference on Artificial Intelligence}, volume~39, pages 5703--5711.

\bibitem[{Liu et~al.(2024)Liu, Qin, Chen, Cheng, and Yang}]{liu2024causality}
Yu~Liu, Guihe Qin, Haipeng Chen, Zhiyong Cheng, and Xun Yang. 2024.
\newblock Causality-inspired invariant representation learning for text-based person retrieval.
\newblock In \emph{Proceedings of the AAAI Conference on Artificial Intelligence}, volume~38, pages 14052--14060.

\bibitem[{Luo et~al.(2019)Luo, Gu, Liao, Lai, and Jiang}]{luo2019bag}
Hao Luo, Youzhi Gu, Xingyu Liao, Shenqi Lai, and Wei Jiang. 2019.
\newblock Bag of tricks and a strong baseline for deep person re-identification.
\newblock In \emph{Proceedings of the IEEE/CVF conference on computer vision and pattern recognition workshops}, pages 0--0.

\bibitem[{Lv et~al.(2024)Lv, Li, Leng, and Gao}]{lv2024mgrl}
Tianle Lv, Shuang Li, Jiaxu Leng, and Xinbo Gao. 2024.
\newblock Mgrl: Mutual-guidance representation learning for text-to-image person retrieval.
\newblock In \emph{ICASSP 2024-2024 IEEE International Conference on Acoustics, Speech and Signal Processing (ICASSP)}, pages 2895--2899. IEEE.

\bibitem[{Ma et~al.(2023)Ma, Sun, Ji, Jiang, Zhuang, and Ji}]{ma2023beat}
Yiwei Ma, Xiaoshuai Sun, Jiayi Ji, Guannan Jiang, Weilin Zhuang, and Rongrong Ji. 2023.
\newblock Beat: Bi-directional one-to-many embedding alignment for text-based person retrieval.
\newblock In \emph{Proceedings of the 31st ACM International Conference on Multimedia}, pages 4157--4168.

\bibitem[{Minderer et~al.(2023)Minderer, Gritsenko, and Houlsby}]{minderer2023scaling}
Matthias Minderer, Alexey Gritsenko, and Neil Houlsby. 2023.
\newblock Scaling open-vocabulary object detection.
\newblock \emph{Advances in Neural Information Processing Systems}, 36:72983--73007.

\bibitem[{Park et~al.(2024)Park, Kim, Jeong, and Kwak}]{park2024plottextbasedpersonsearch}
Jicheol Park, Dongwon Kim, Boseung Jeong, and Suha Kwak. 2024.
\newblock \href {https://arxiv.org/abs/2409.13475} {Plot: Text-based person search with part slot attention for corresponding part discovery}.
\newblock \emph{Preprint}, arXiv:2409.13475.

\bibitem[{Qin et~al.(2024)Qin, Chen, Peng, Peng, Zhou, and Hu}]{qin2024noisy}
Yang Qin, Yingke Chen, Dezhong Peng, Xi~Peng, Joey~Tianyi Zhou, and Peng Hu. 2024.
\newblock Noisy-correspondence learning for text-to-image person re-identification.
\newblock In \emph{IEEE International Conference on Computer Vision and Pattern Recognition (CVPR)}.

\bibitem[{Qin et~al.(2022)Qin, Peng, Peng, Wang, and Hu}]{qin2022deepdecl}
Yang Qin, Dezhong Peng, Xi~Peng, Xu~Wang, and Peng Hu. 2022.
\newblock Deep evidential learning with noisy correspondence for cross-modal retrieval.
\newblock In \emph{Proceedings of the 30th ACM International Conference on Multimedia}, pages 4948--4956.

\bibitem[{Radford et~al.(2021{\natexlab{a}})Radford, Kim, Hallacy, Ramesh, Goh, Agarwal, Sastry, Askell, Mishkin, Clark et~al.}]{clip}
Alec Radford, Jong~Wook Kim, Chris Hallacy, Aditya Ramesh, Gabriel Goh, Sandhini Agarwal, Girish Sastry, Amanda Askell, Pamela Mishkin, Jack Clark, et~al. 2021{\natexlab{a}}.
\newblock Learning transferable visual models from natural language supervision.
\newblock In \emph{International conference on machine learning}, pages 8748--8763. PMLR.

\bibitem[{Radford et~al.(2021{\natexlab{b}})Radford, Kim, Hallacy, Ramesh, Goh, Agarwal, Sastry, Askell, Mishkin, Clark et~al.}]{radford2021learning}
Alec Radford, Jong~Wook Kim, Chris Hallacy, Aditya Ramesh, Gabriel Goh, Sandhini Agarwal, Girish Sastry, Amanda Askell, Pamela Mishkin, Jack Clark, et~al. 2021{\natexlab{b}}.
\newblock Learning transferable visual models from natural language supervision.
\newblock In \emph{ICML}, pages 8748--8763. PMLR.

\bibitem[{Shao et~al.(2023)Shao, Zhang, Ding, Wang, and Wang}]{shao2023unified}
Zhiyin Shao, Xinyu Zhang, Changxing Ding, Jian Wang, and Jingdong Wang. 2023.
\newblock Unified pre-training with pseudo texts for text-to-image person re-identification.
\newblock In \emph{Proceedings of the IEEE/CVF International Conference on Computer Vision}, pages 11174--11184.

\bibitem[{Shao et~al.(2022)Shao, Zhang, Fang, Lin, Wang, and Ding}]{shao2022learning}
Zhiyin Shao, Xinyu Zhang, Meng Fang, Zhifeng Lin, Jian Wang, and Changxing Ding. 2022.
\newblock Learning granularity-unified representations for text-to-image person re-identification.
\newblock In \emph{ACM MM}, pages 5566--5574.

\bibitem[{Shen et~al.(2023)Shen, Shu, Du, and Tang}]{shen2023pedestrian}
Fei Shen, Xiangbo Shu, Xiaoyu Du, and Jinhui Tang. 2023.
\newblock Pedestrian-specific bipartite-aware similarity learning for text-based person retrieval.
\newblock In \emph{Proceedings of the 31st ACM International Conference on Multimedia}, pages 8922--8931.

\bibitem[{Shu et~al.(2022)Shu, Wen, Wu, Chen, Song, Qiao, Ren, and Wang}]{shu2022see}
Xiujun Shu, Wei Wen, Haoqian Wu, Keyu Chen, Yiran Song, Ruizhi Qiao, Bo~Ren, and Xiao Wang. 2022.
\newblock See finer, see more: Implicit modality alignment for text-based person retrieval.
\newblock In \emph{European Conference on Computer Vision}, pages 624--641. Springer.

\bibitem[{Shu et~al.(2026)Shu, Liu, and Xie}]{shu2026agenticretrievalaugmentedgenerationfinancial}
Yang Shu, Yingmin Liu, and Zequn Xie. 2026.
\newblock \href {https://arxiv.org/abs/2605.05409} {Agentic retrieval-augmented generation for financial document question answering}.
\newblock \emph{Preprint}, arXiv:2605.05409.

\bibitem[{Song et~al.(2024)Song, Hu, and Zhao}]{song2024diverse}
Zifan Song, Guosheng Hu, and Cairong Zhao. 2024.
\newblock Diverse person: Customize your own dataset for text-based person search.
\newblock In \emph{Proceedings of the AAAI Conference on Artificial Intelligence}, volume~38, pages 4943--4951.

\bibitem[{Wang et~al.(2021)Wang, Luo, Lin, and Li}]{wang2021text}
Chengji Wang, Zhiming Luo, Yaojin Lin, and Shaozi Li. 2021.
\newblock Text-based person search via multi-granularity embedding learning.
\newblock In \emph{IJCAI}, pages 1068--1074.

\bibitem[{Wang et~al.(2020)Wang, Fang, Wang, and Yang}]{wang2020vitaa}
Zhe Wang, Zhiyuan Fang, Jun Wang, and Yezhou Yang. 2020.
\newblock Vitaa: Visual-textual attributes alignment in person search by natural language.
\newblock In \emph{Computer Vision--ECCV 2020: 16th European Conference, Glasgow, UK, August 23--28, 2020, Proceedings, Part XII 16}, pages 402--420. Springer.

\bibitem[{Wang et~al.(2022)Wang, Zhu, Xue, Wan, Liu, Wang, and Li}]{wang2022look}
Zijie Wang, Aichun Zhu, Jingyi Xue, Xili Wan, Chao Liu, Tian Wang, and Yifeng Li. 2022.
\newblock Look before you leap: Improving text-based person retrieval by learning a consistent cross-modal common manifold.
\newblock In \emph{ACM MM}, pages 1984--1992.

\bibitem[{Wu et~al.(2021)Wu, Yan, Han, Li, Zou, and Cui}]{wu2021lapscore}
Yushuang Wu, Zizheng Yan, Xiaoguang Han, Guanbin Li, Changqing Zou, and Shuguang Cui. 2021.
\newblock Lapscore: language-guided person search via color reasoning.
\newblock In \emph{Proceedings of the IEEE/CVF International Conference on Computer Vision}, pages 1624--1633.

\bibitem[{Xie(2026)}]{xie2026conquer}
Zequn Xie. 2026.
\newblock Conquer: Context-aware representation with query enhancement for text-based person search.
\newblock \emph{arXiv preprint arXiv:2601.18625}.

\bibitem[{Xie et~al.(2025{\natexlab{a}})Xie, Ji, and Meng}]{xie2025dynamic}
Zequn Xie, Haoming Ji, and Lingwei Meng. 2025{\natexlab{a}}.
\newblock Dynamic uncertainty learning with noisy correspondence for text-based person search.
\newblock \emph{arXiv preprint arXiv:2505.06566}.

\bibitem[{Xie et~al.(2026{\natexlab{a}})Xie, Liu, Zhang, Lin, Cai, and Jin}]{xie2026hvd}
Zequn Xie, Xin Liu, Boyun Zhang, Yuxiao Lin, Sihang Cai, and Tao Jin. 2026{\natexlab{a}}.
\newblock Hvd: Human vision-driven video representation learning for text-video retrieval.
\newblock \emph{arXiv preprint arXiv:2601.16155}.

\bibitem[{Xie et~al.(2026{\natexlab{b}})Xie, Luo, Wang, Cai, Jin, Zhao, and Tang}]{xie2026bridgingposesemanticgapcascade}
Zequn Xie, Guijin Luo, Chuxin Wang, Sihang Cai, Tao Jin, Zhou Zhao, and Yixuan Tang. 2026{\natexlab{b}}.
\newblock \href {https://arxiv.org/abs/2604.23282} {Bridging the pose-semantic gap: A cascade framework for text-based person anomaly search}.
\newblock \emph{Preprint}, arXiv:2604.23282.

\bibitem[{Xie et~al.(2025{\natexlab{b}})Xie, Wang, Wang, Cai, Wang, and Jin}]{xie2025chat}
Zequn Xie, Chuxin Wang, Yeqiang Wang, Sihang Cai, Shulei Wang, and Tao Jin. 2025{\natexlab{b}}.
\newblock Chat-driven text generation and interaction for person retrieval.
\newblock In \emph{Proceedings of the 2025 Conference on Empirical Methods in Natural Language Processing}, pages 5259--5270.

\bibitem[{Xie et~al.(2026{\natexlab{c}})Xie, Zhang, Lin, and Jin}]{xie2026delving}
Zequn Xie, Boyun Zhang, Yuxiao Lin, and Tao Jin. 2026{\natexlab{c}}.
\newblock Delving deeper: Hierarchical visual perception for robust video-text retrieval.
\newblock \emph{arXiv preprint arXiv:2601.12768}.

\bibitem[{Yan et~al.(2023{\natexlab{a}})Yan, Dong, Liu, Zhang, and Tang}]{yan2023learning}
Shuanglin Yan, Neng Dong, Jun Liu, Liyan Zhang, and Jinhui Tang. 2023{\natexlab{a}}.
\newblock Learning comprehensive representations with richer self for text-to-image person re-identification.
\newblock In \emph{Proceedings of the 31st ACM international conference on multimedia}, pages 6202--6211.

\bibitem[{Yan et~al.(2023{\natexlab{b}})Yan, Dong, Zhang, and Tang}]{yan2023clip}
Shuanglin Yan, Neng Dong, Liyan Zhang, and Jinhui Tang. 2023{\natexlab{b}}.
\newblock Clip-driven fine-grained text-image person re-identification.
\newblock \emph{IEEE Transactions on Image Processing}.

\bibitem[{Yan et~al.(2022)Yan, Tang, Zhang, and Tang}]{yan2022image}
Shuanglin Yan, Hao Tang, Liyan Zhang, and Jinhui Tang. 2022.
\newblock Image-specific information suppression and implicit local alignment for text-based person search.
\newblock \emph{arXiv preprint arXiv:2208.14365}.

\bibitem[{Yang et~al.(2023)Yang, Zhou, Wang, Wu, Zhu, and Zheng}]{yang2023towards}
Shuyu Yang, Yinan Zhou, Yaxiong Wang, Yujiao Wu, Li~Zhu, and Zhedong Zheng. 2023.
\newblock Towards unified text-based person retrieval: A large-scale multi-attribute and language search benchmark.
\newblock In \emph{Proceedings of the 2023 {ACM} on Multimedia Conference}.

\bibitem[{Zhang and Lu(2018)}]{zhang2018deep}
Ying Zhang and Huchuan Lu. 2018.
\newblock Deep cross-modal projection learning for image-text matching.
\newblock In \emph{Proceedings of the European conference on computer vision (ECCV)}, pages 686--701.

\bibitem[{Zhao et~al.(2024)Zhao, Liu, Lu, Chu, and Yu}]{zhao2024unifying}
Zhiwei Zhao, Bin Liu, Yan Lu, Qi~Chu, and Nenghai Yu. 2024.
\newblock Unifying multi-modal uncertainty modeling and semantic alignment for text-to-image person re-identification.
\newblock In \emph{Proceedings of the AAAI Conference on Artificial Intelligence}, volume~38, pages 7534--7542.

\bibitem[{Zheng et~al.(2020)Zheng, Zheng, Garrett, Yang, Xu, and Shen}]{zheng2020dual}
Zhedong Zheng, Liang Zheng, Michael Garrett, Yi~Yang, Mingliang Xu, and Yi-Dong Shen. 2020.
\newblock Dual-path convolutional image-text embeddings with instance loss.
\newblock \emph{ACM Transactions on Multimedia Computing, Communications, and Applications}, 16(2):1--23.

\bibitem[{Zhu et~al.(2021)Zhu, Wang, Li, Wan, Jin, Wang, Hu, and Hua}]{zhu2021dssl}
Aichun Zhu, Zijie Wang, Yifeng Li, Xili Wan, Jing Jin, Tian Wang, Fangqiang Hu, and Gang Hua. 2021.
\newblock Dssl: Deep surroundings-person separation learning for text-based person retrieval.
\newblock In \emph{Proceedings of the 29th ACM International Conference on Multimedia}, pages 209--217.

\bibitem[{Zuo et~al.(2024)Zuo, Zhou, Nie, Zhang, Guo, Sang, Wang, and Gao}]{zuo2024ufinebench}
Jialong Zuo, Hanyu Zhou, Ying Nie, Feng Zhang, Tianyu Guo, Nong Sang, Yunhe Wang, and Changxin Gao. 2024.
\newblock Ufinebench: Towards text-based person retrieval with ultra-fine granularity.
\newblock In \emph{Proceedings of the IEEE/CVF Conference on Computer Vision and Pattern Recognition}, pages 22010--22019.

\end{thebibliography}

\clearpage
\appendix
\clearpage

\section{Supplementary Material}

This supplementary material provides additional details regarding our experimental setup, the P-VLG benchmark statistics, and qualitative examples to further illustrate the capabilities of our proposed method.

\subsection{Implementation Details}
\label{sec:supp_implementation}

\paragraph{Model and Training}
Our framework, ROGLE, is built upon the CLIP ViT-B/16 architecture for both the image and text encoders, initialized with official pre-trained weights. The model is trained for 60 epochs using the Adam optimizer with a batch size of 64 and a weight decay of $5 \times 10^{-4}$. We employ a two-stage learning rate schedule: a linear warm-up for the first 5 epochs from $1 \times 10^{-6}$ to $1 \times 10^{-5}$, followed by a cosine decay schedule. The temperature parameter $\tau$ in the contrastive loss is set to 0.02. All experiments were conducted on a single NVIDIA RTX 4090 GPU.

\paragraph{Data Processing and RSM}
For data augmentation, input images are resized to $384 \times 128$ and subjected to random horizontal flipping, cropping, and erasing. Textual captions are tokenized using CLIP's BPE tokenizer with a maximum sequence length of 77. For the Region-to-Sentence Matching (RSM) module, we use the Segment Anything Model (SAM) with a ViT-H backbone to generate region proposals in automatic mode. Long-form captions are decomposed into atomic sentences using the spaCy library.

\subsection{P-VLG Benchmark Details}
\label{sec:supp_benchmark}

The P-VLG benchmark was created to address the need for datasets with longer, more descriptive captions and fine-grained annotations for text-based person search.P-VLG provides a significantly larger number of captions and contains region-level alignments, making it a valuable resource for developing and evaluating models with fine-grained understanding capabilities.

\subsection{Qualitative Examples}
\label{sec:supp_qualitative}

\begin{figure*}[h]
    \centering
    % Placeholder for your actual images. 
    % Please ensure the paths 'figs/ex1.pdf', etc., are correct in your project.
    \includegraphics[width=\textwidth]{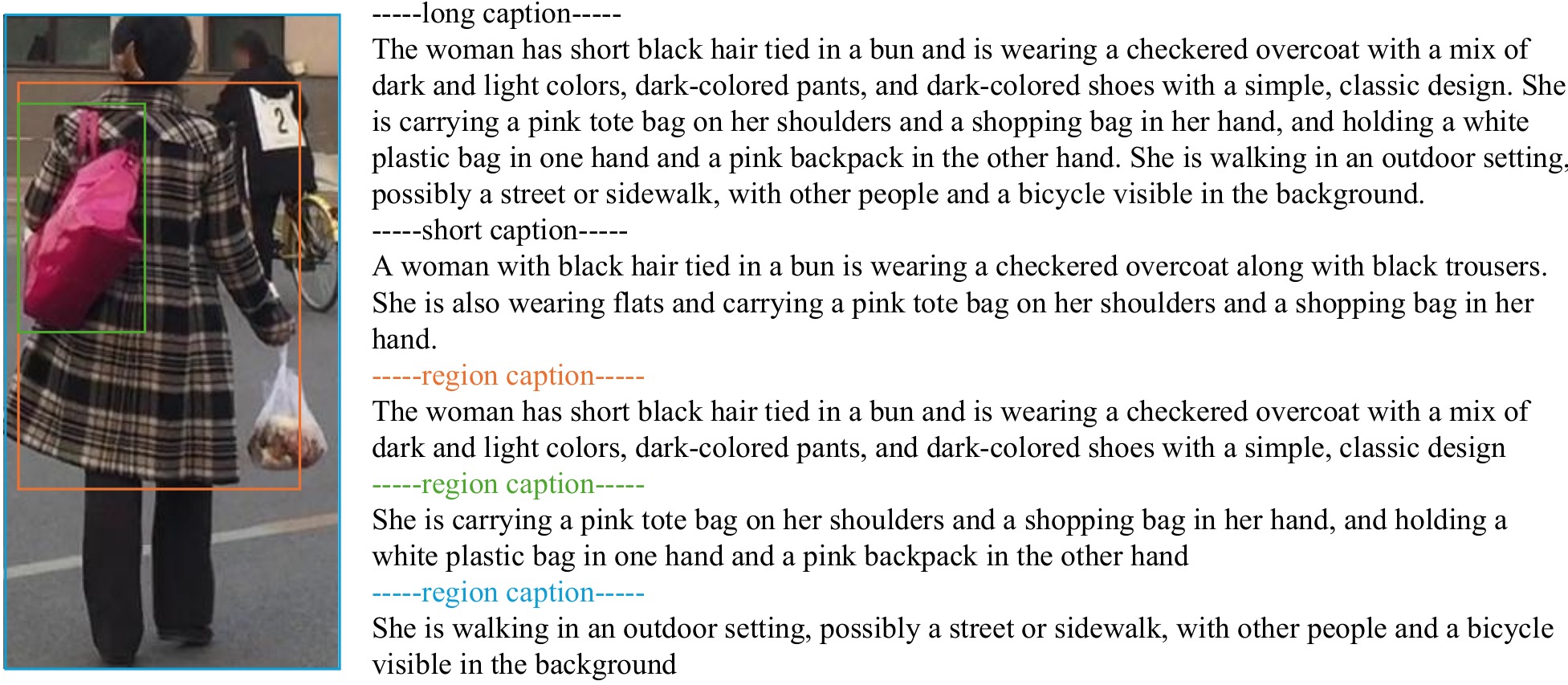}
    \includegraphics[width=0.49\textwidth]{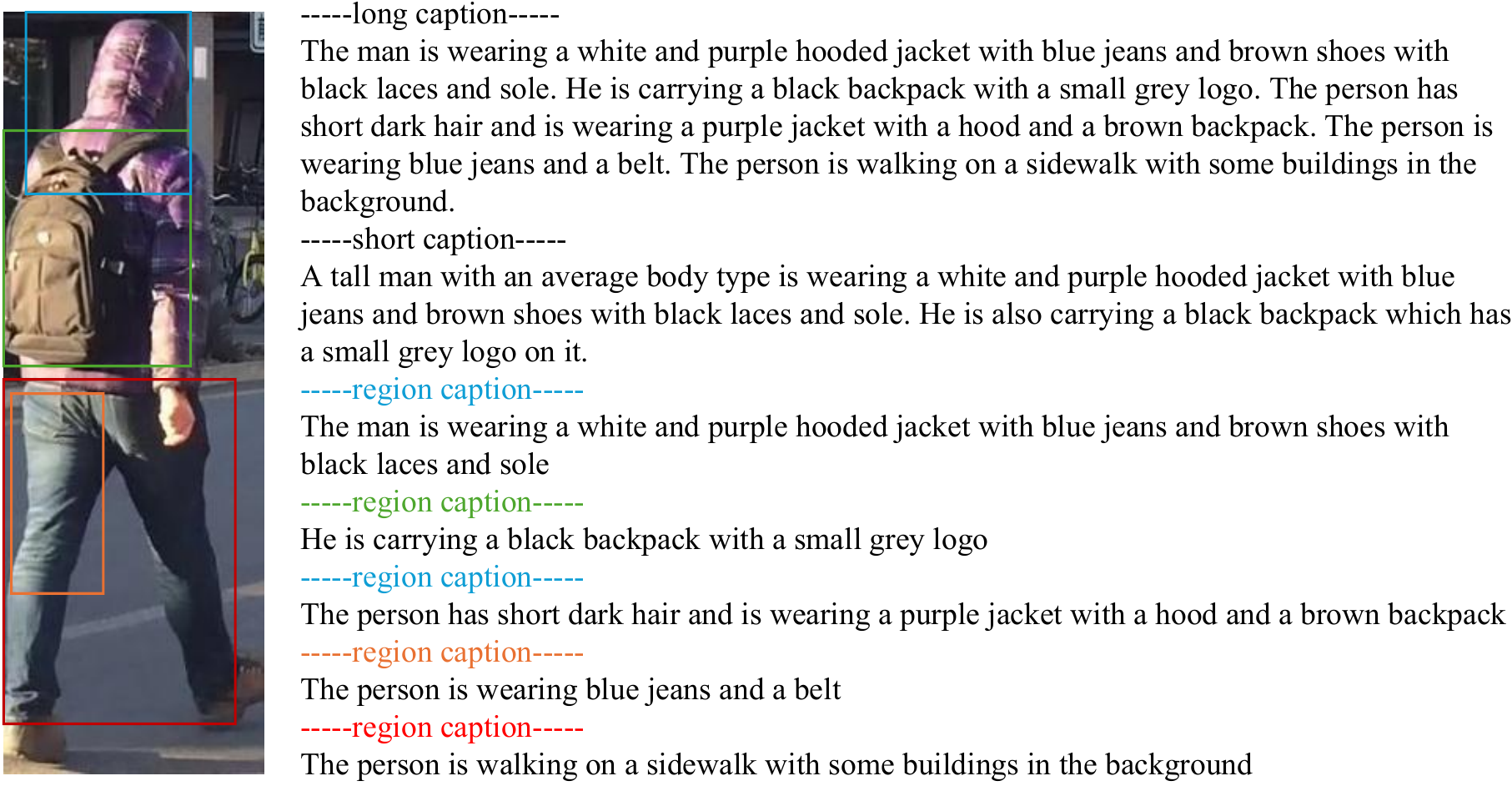}\hfill
    \includegraphics[width=0.49\textwidth]{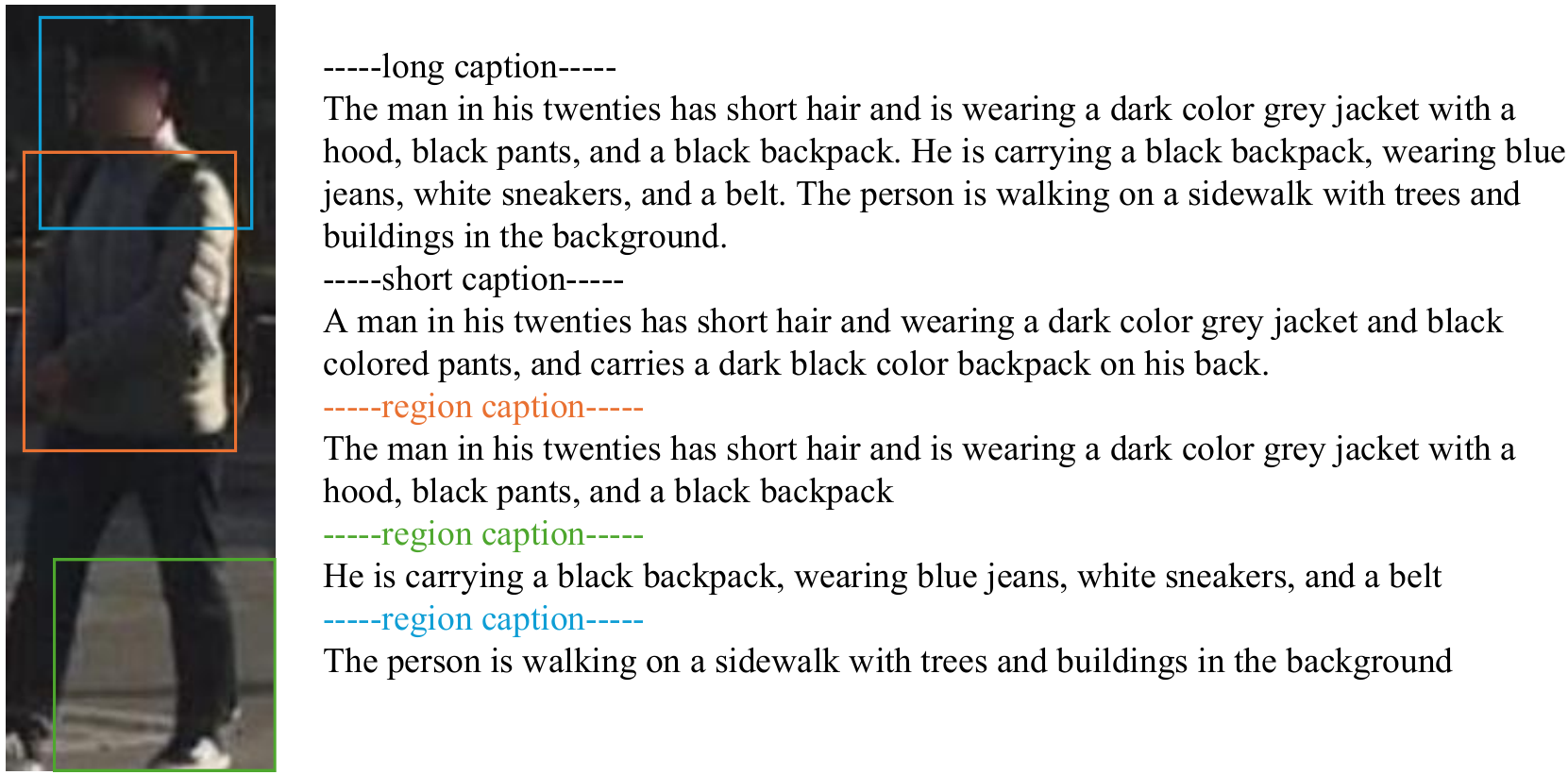}
    \caption{Qualitative examples from the P-VLG benchmark, illustrating the fine-grained alignment between textual descriptions and their corresponding image regions generated by our automated RSM pipeline. The long caption (top of each example) is successfully decomposed into semantically coherent region-sentence pairs (bottom).}
    \label{fig:supp_examples}
\end{figure*}

To visually demonstrate the effectiveness of our Region-to-Sentence Matching (RSM) pipeline, Figure~\ref{fig:supp_examples} presents qualitative examples from the P-VLG test set. These examples show how a single, long-form caption is automatically decomposed and aligned with specific, semantically relevant regions of the pedestrian image.

For instance, in the first example (top), the global caption describes a woman with a "checkered overcoat," a "pink tote bag," and her action of "walking in an outdoor setting". Our RSM module successfully generates three distinct region-sentence pairs: one aligning the description of the overcoat and clothing with a bounding box around her torso, a second aligning the description of the bags she is carrying with a box around her arms and bags, and a third linking the description of the background scene to the overall image context. This automated process provides high-quality, fine-grained supervision that is crucial for teaching the model to ground textual phrases in corresponding visual evidence.

\end{document}